\documentclass[conference]{IEEEtran}
\IEEEoverridecommandlockouts
\usepackage{cite}
\usepackage{amsmath,amssymb,amsfonts}
\usepackage{algorithmic}
\usepackage{graphicx}
\usepackage{textcomp}
\usepackage{xcolor}
\usepackage{adjustbox}
\usepackage{hyperref} 
\usepackage{multirow} 
\def\BibTeX{{\rm B\kern-.05em{\sc i\kern-.025em b}\kern-.08em
    T\kern-.1667em\lower.7ex\hbox{E}\kern-.125emX}}

\begin{document}

 \title{SciClaimHunt: A Large Dataset for Evidence-based Scientific Claim Verification}

\author{
  Sujit Kumar$^{1*}$, Anshul Sharma$^{2*}$, Siddharth Hemant Khincha$^{1*}$\thanks{* Equal contributions}, 
    Gargi Shroff$^{2}$, Sanasam Ranbir Singh$^{1}$, Rahul Mishra$^{2}$\\
    $^{1}$Department of Computer Science and Engineering, \\
    Indian Institute of Technology, Guwahati, Assam, India \\
    $^{2}$ International Institute of Information Technology Hyderabad\\
    \texttt{\{sujitkumar,s.khincha\}@alumni.iitg.ac.in}, \texttt{ranbir@iitg.ac.in}\\
    \texttt{\{anshul.s,gargi.shroff\}@research.iiit.ac.in}, \texttt{rahul.mishra@iiit.ac.in}
    
}

\maketitle

\begin{abstract}
Verifying scientific claims presents a significantly greater challenge than verifying political or news-related claims. Unlike the relatively broad audience for political claims, the users of scientific claim verification systems can vary widely, ranging from researchers testing specific hypotheses to everyday users seeking information on a medication. Additionally, the evidence for scientific claims is often highly complex, involving technical terminology and intricate domain-specific concepts that require specialized models for accurate verification. Despite considerable interest from the research community, there is a noticeable lack of large-scale scientific claim verification datasets to benchmark and train effective models. To bridge this gap, we introduce two large-scale datasets, \textit{SciClaimHunt} and  \textit{SciClaimHunt\_Num}, derived from scientific research papers. We propose several baseline models tailored for scientific claim verification to assess the effectiveness of these datasets. Additionally, we evaluate models trained on \textit{SciClaimHunt} and  \textit{SciClaimHunt\_Num} against existing scientific claim verification datasets to gauge their quality and reliability. Furthermore, we conduct human evaluations of the dataset’s claims and perform error analysis to assess the effectiveness of the proposed baseline models. Our findings indicate that\textit{SciClaimHunt} and  \textit{SciClaimHunt\_Num} serve as highly reliable resources for training models in scientific claim verification.

\end{abstract}


\section{Introduction}

Fact-checking plays a vital role in information validation, serving as a crucial defense against the widespread dissemination of misinformation and disinformation. Fact-checking is the process of verifying the authenticity of a claim with the help of evidence documents that either support or refute the claim, where a claim is defined as a factual statement subjected to  verification~\cite{vlachos2014fact,thorne2018fever,hanselowski2019richly,thorne2019fever2, setty2020deep}. Initial studies~\cite{derczynski2017semeval,gorrell-etal-2019-semeval,barron2020overview,elsayed2019checkthat,nakov2018overview,hassan2017claimbuster, SADHAN_mishra,mishra-etal-2020-generating} on fact-checking primarily concentrated on curating datasets and developing methodologies for verifying claims related to politics, current affairs, 
newscast and discussions on social media. Scientific claims are grounded in empirical data derived from scientific studies and reports, whereas ordinary claims often express opinions or assertions about potential benefits or courses of action.  However, verifying scientific claims poses unique challenges in dataset curation and model development. Political and general claims can be easily fact-checked by general annotators or journalists, with numerous fact-checking websites available for such claims. In contrast, dedicated fact-checking platforms for scientific claims are scarce, and verifying these claims requires annotators with substantial domain expertise ~\cite{wadden2020fact}. Given these unique challenges in scientific claim verification, the study \cite{wadden2020fact} proposes a scientific claim verification dataset~\textit{SCIFACT} by manually constructing scientific claims derived from research paper abstracts, paired with abstracts of research papers that either support or refute the corresponding claim. Similarly, study~\cite{wadden2022scifact} introduces \textit{SCIFACT-OPEN} dataset for scientific claim verification by compiling pairs of scientific claims and abstracts of multiple research papers that either support or refute the scientific claim

\begin{figure}
    \centering
    \includegraphics[width=2in]{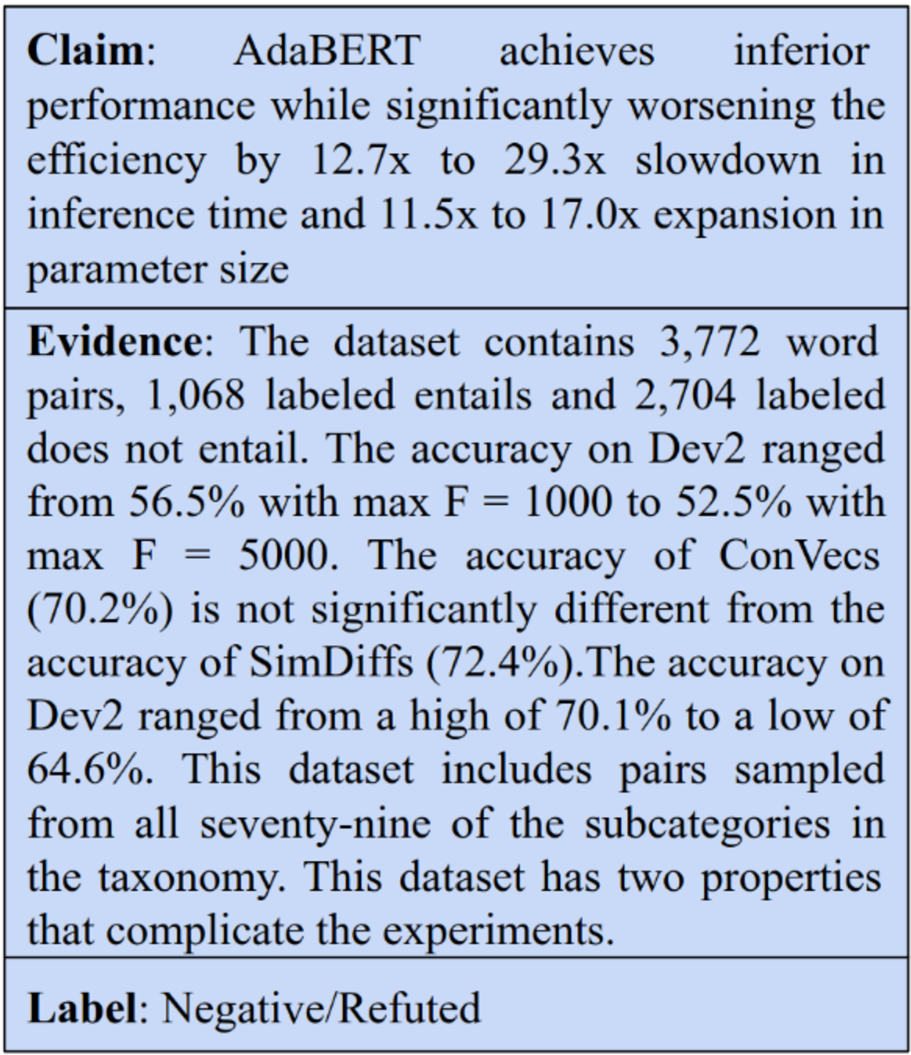}
    \vspace{\baselineskip} 
    \includegraphics[width=1.3in,height=2.3in]{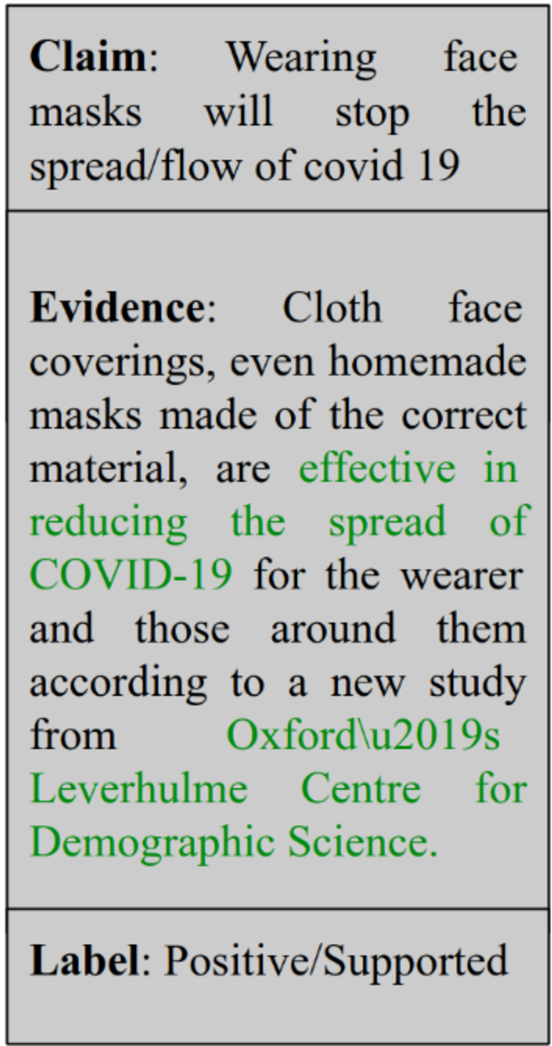}
    \caption{The left sub-figure shows an example of a negative scientific claim involving numerals and cardinal numbers from \textit{SciClaimHunt\_Num}, refuted by evidence extracted from a research paper. The right sub-figure showcases a positive scientific claim from the \textit{SciClaimHunt} dataset, supported by evidence retrieved from the research paper.}
    \label{fig:main_example}
\end{figure}

Further, the work~\cite{wright2022generating} proposes two methods for curating scientific claim verification datasets in the biomedical domain, namely \textit{Knowledge Base Informed Negations} and \textit{CLAIMGEN-BART}. The CLAIMGEN-BART method initially extracts the citation, along with the preceding and subsequent sentences, from a given research paper. Next, they encode these extracted sentences using the BART encoder and pass them to the BART decoder to generate a scientific claim supported by the research paper. In contrast, the \textit{Knowledge Base Informed Negations} method replaces the named entity in a scientific claim, which is supported by the research paper to generate a scientific claim that is refuted by the research paper. However, the existing datasets for scientific claim verification exhibit several notable limitations. (i) The existing datasets in the literature are limited in size, typically containing only a few thousand samples, which is insufficient for training generic models for scientific claim detection across diverse categories and domains\cite{vladika2023scientific}. (ii) Existing datasets for scientific claim verification predominantly utilize research paper abstracts as supporting or refuting evidence. However, due to their concise nature, abstracts often lack the comprehensive detail necessary to provide sufficient evidence for robust claim verification. (iii) Existing datasets either rely on manually extracted claims or derive them from research paper references, overlooking claims and observations found in the results, discussion, and conclusion sections. However, these sections often contain the most critical claims and insights related to scientific studies. (iv) Absence of datasets in literature for scientific claims with cardinal or numeral values. \par

Motivated by such limitations with the existing datasets for scientific claim verification. This study proposes two novel datasets for scientific claim verification tasks: \textbf{SciCliamHunt} and~\textbf{SciClaimHunt\_Num}. Our proposed SciClaimHunt dataset leverages few-shot prompting with Large Language Models (LLMs) to generate scientific claims grounded in supporting scientific documents. SciClaimHunt generates refuted claims using two methods: (i) negating a scientific claim supported by scientific evidence and (ii) named entity replacement within a scientific claim supported by scientific evidence. We also curate \textbf{SciClaimHunt\_Num} dataset, a subset of \textit{SciClaimHunt} dataset, dedicated to scientific claims involving numerical or cardinal values, where a model also needs to verify the consistency of numerals and cardinal numbers within the scientific claims and scientific evidence, along with the consistency and contextual similarity between the scientific claim and scientific evidence, to decide whether a scientific claim is supported or refuted by the scientific evidence. Figure~\ref{fig:main_example} presents the examples of \textit{SciCliamHunt} and~\textit{ SciClaimHunt\_Num} datasets. We also propose non-trivial and suitably motivated baseline methods to evaluate the effectiveness of our proposed datasets for scientific claim verification. We also assess the quality and reliability of our proposed datasets through various ablation studies, human annotation evaluations, and error analysis. The quality and reliability evaluation of proposed datasets suggests that our proposed datasets \textit{SciCliamHunt} and \textit{ SciClaimHunt\_Num} are reliable and effective in training models for scientific claim validations.

\section{Related work}
In the literature, studies\cite{guo2022survey,vladika2023scientific} briefly review and analyze the datasets and methodologies related to fact-checking. In litrature several datasets have been proposed to support general fact-checking research across various domains, including politics, journalism, and social media~\cite{vlachos2014fact,popat2017truth,thorne2018fever,thorne2018fact,alhindi2018your,chen2019seeing,hanselowski2019richly,augenstein2019multifc,atanasova2024multi,sathe2020automated,saakyan2021covid,schuster2021get,aly2021feverous}.
Scientific fact-checking, a subfield of fact-checking, focuses on verifying claims about scientific knowledge, thereby combating misinformation while also supporting scientific inquiry, interpretation and public understanding of research~\cite{vladika2023scientific}. Initial  study~\cite{wadden2020fact} on scientific fact-checking proposed the \textit{SCIFACT} dataset for scientific claim verification, using citance as claims and research abstracts as evidence. In contrast, \textit{SCIFACT-OPEN}\cite{wadden2022scifact} extends the study~\cite{wadden2020fact} by collecting abstracts from multiple papers to support or refute the claims. Further study~\cite{wright2022generating} 
introduced an encoder-decoder model that utilizes sentences surrounding stances and BART to generate claims supported by scientific evidence, with negative claims through named entity replacements to generate the claim refuted for scientific evidence. However, existing scientific claim verification datasets in the litrature have several key limitations. A primary concern is their limited size, as they typically consist of only a few hundred or thousand samples. This is inadequate for training models that can generalize effectively across diverse categories and domains \cite{vladika2023scientific}. Moreover, these datasets primarily depend on research paper abstracts as external evidence. However, abstracts offer only a concise summary and often lack the comprehensive details necessary for rigorous scientific claim verification \cite{vladika2023scientific}. Another limitation is the emphasis on extracting claims solely from sentences containing citations of research papers, which overlooks significant claims and insights presented in the results, discussion, and conclusion sections. Furthermore, there is the absence of a scientific claim verification dataset in the literature, which contains scientific claims and scientific evidence involving cardinal or numeral values to train models for numeral-aware scientific claim verifications. Motivated by such limitations with existing datasets for scientific claim verification in literature, this study proposes two novel datasets for scientific claim verification tasks \textit{SciCliamHunt} and \textit{ SciClaimHunt\_Num}.


\section{Proposed Datasets}
\label{sec:dataset}
A claim is considered valid if it is fluent, atomic, decontextualized, and precisely conveys the meaning of the source sentence\cite{wright2022generating}.  Given a scientific research paper $\mathcal{R}$ that contains results, discussions, and a conclusion, this study aims to generate a scientific claim $\mathcal{C}$ from sentences in the discussion and conclusion section of the research paper of $\mathcal{R}$, which is supported or refuted by the content of the research paper $\mathcal{R}$. A scientific claim $\mathcal{C}$ is labelled as \textit{Positive} if a scientific research paper supports the claim $\mathcal{C}$, while $\mathcal{C}$ is considered \textit{Negative} if a scientific research paper refutes claim $\mathcal{C}$. This study utilizes the research paper corpus reported in study~ ~\cite{moosavi2021scigen} to construct the \textit{SciClaimHunt} and \textit{SciClaimHunt\_Num} datasets. Ethical considerations related to the use of this corpus are discussed in Section~\ref{sec:ethics}. 


\subsection{Positive Claim Generation}
\label{subsec:pos}
Initially, we form a subset $\mathcal{F}$ from the research paper corpus $\mathcal{D}$ by randomly selecting twelve papers from $\mathcal{D}$. Next, from each research paper $\mathcal{R} \in \mathcal{F}$, we manually extract a scientific claim  $\mathcal{C}$, which is supported by the content of research paper $\mathcal{R}$. Subsequently, inspired by the \textit{Promptagator} methods outlined in the study\cite{daipromptagator}, we propose {\it Promp\_Claim\_Generator} using a few-shot prompting approach for generating positive claims. Subsequently, considering twelve pairs of manually extracted claims and corresponding paragraphs from which the claim is extracted, our proposed {\it Promp\_Claim\_Generator} method for generating positive claims utilizes instruction prompts $\mathcal{I}$ as follows.  
\begin{equation}
 \mathcal{I} \text{=}   \Big( (\mathcal{P}_i, \mathcal{C}_i), ... , (\mathcal{P}_k, \mathcal{C}_k), (\mathcal{P}_{new} ) \Big)
\end{equation}
Where $\mathcal{P}_i, \mathcal{C}_i ; \forall ; i = 1 ; \text{to} ; k$ are the $k$ pairs of manually extracted claims $\mathcal{C}_i$ and their corresponding source paragraphs $\mathcal{P}_i$, from which the claims is extracted. These $k$  pair of $(\mathcal{P}_I, \mathcal{C}_i)$ are used as $k$ shot examples to LLMs, and the value of $k$ is twelve. Meanwhile, $\mathcal{P}_{new}$ represents the news paragraphs from which a positive claim needs to be generated. Here, $\mathcal{P}_{new} \in \mathcal{R}$ denotes a new source paragraph from a research paper $\mathcal{R} \in \mathcal{D}$ within the research paper corpus $\mathcal{D}$. We feed the instruction prompts $\mathcal{I}$ to Llama\cite{touvron2023llama}~\footnote{\href{https://huggingface.co/TheBloke/Llama-2-13B-chat-GGML}{TheBloke/Llama-2-13B-chat-GGML}} to generate the new claim over paragraph $\mathcal{P}_{new}$ from research paper $\mathcal{R} \in \mathcal{D}$. By running  Llama\cite{touvron2023llama} with instruction prompts $\mathcal{I}$ on paragraphs extracted from result and discussion and conclusion sections of each research paper $\mathcal{R} \in \mathcal{D}$, we generate a large collection of positive claims $\mathcal{C}$ which supported by contents of respective research paper $\mathcal{R}$. 
  
\subsection{Negative Claim Generation}
\label{subsec:neg}
This study considers four different methods to generate and extract negative claims from the results, discussion, and conclusion sections of scientific research papers, refuted by the contents within the same research paper as follows: (i) We adopt few-shot prompting with Llama\cite{touvron2023llama}, where prompt instructions $\mathcal{P}_r$ and eight examples are provided to Llama along with the positive claim extracted in subsection~\ref{subsec:pos} to generate a negate of the positive claim as a negative claim, effectively conveying the opposite meaning of the positive claim. Our prompt instructions $\mathcal{P}_r$ to Llama to generate a negative claim is as follows: \textit{I want you to negate claims extracted from a given scientific paragraph. The negations must be strong negations that are surely false. Examples may be changing results and methodology mentioned in the claims.} (ii) We generate negative claims by mismatching positive claims with different research papers. Specifically, given a true positive claim and research paper pair $(\mathcal{C}_i,\mathcal{R}_i)$, we create a negative pair by mismatching the claim from the $i^{th}$ scientific document with the $j^{th}$ scientific document, resulting in a negative pair of $(\mathcal{C}_i,\mathcal{R}_j)$, where $\mathcal{C}_i$ is refuted by the contents of $\mathcal{R}_j$. (iii) We also generate negative claims by altering the cardinal and numeral values in the positive claims, resulting in claims with incorrect numerical information, which is refuted by the contents of research papers. (iii) We also generate negative claims by altering the cardinal and numeral values in the positive claims, resulting in claims with incorrect numerical information, which is refuted by the contents of research papers. (iv) We also generate negative claims by following the named entities in the positive claim, following the \textit{Knowledge Base Informed Negations} approach as reported in the study\cite{wright2022generating}.

\begin{table}[t]
\centering
\caption{Characteristics of Experimental Datasets. Here, \#Claim and \#Scientific Paper indicate the average number of words in the Claim and Scientific research paper, respectively. Similarly, \#Sen indicates the average number of sentences in a scientific research paper.}
\label{tab:proposed_dataset}
\renewcommand{\arraystretch}{1.2} 
\setlength{\tabcolsep}{4pt} 
\begin{adjustbox}{max width=\columnwidth}
\begin{tabular}{l l r r r r r r} 
\hline
\textbf{Dataset} & \textbf{Split} & \textbf{Support} & \textbf{Refutes} & \textbf{Total} & \textbf{\#Claim} & \textbf{\#Paper} & \textbf{\#Sen} \\  
\hline
\multirow{3}{*}{\textbf{SciClaimHunt}}  
& Train & 37,597 & 49,512 & 87,109 & 21.1 & 4,497.8 & 265.4 \\  
& Test  & 4,769  & 6,131  & 10,900 & 21.0 & 4,505.1 & 265.8 \\  
& Dev   & 4,678  & 6,206  & 10,884 & 21.0 & 4,495.3 & 264.0 \\  
\multirow{3}{*}{\textbf{SciClaimHunt\_Num}}  
& Train & 9,625  & 10,694 & 20,319 & 25.0 & 4,475.1 & 266.5 \\  
& Test  & 1,115  & 1,343  & 2,458  & 24.8 & 4,470.4 & 265.1 \\  
& Dev   & 1,162  & 1,385  & 2,547  & 24.9 & 4,313.9 & 257.6 \\  
\hline
\end{tabular}
\end{adjustbox}
\vspace{-2mm} 
\end{table}

\begin{table}[t]
\centering
\caption{Inter-annotator scores for claim evaluation.}
\label{tab:annot_score}
\renewcommand{\arraystretch}{1.2} 
\setlength{\tabcolsep}{4pt} 
\begin{adjustbox}{max width=\columnwidth}
\begin{tabular}{l r r r r} 
\hline
\textbf{Metric} & \textbf{Fluency} & \textbf{Atomicity} & \textbf{De-Contextualization} & \textbf{Faithfulness} \\  
\hline
\textbf{Krippendorff} & 0.704 & 0.703 & 0.784 & 0.693 \\  
\textbf{Fleiss kappa} & 0.703 & 0.823 & 0.761 & 0.699 \\  
\hline
\end{tabular}
\end{adjustbox}
\vspace{-2mm} 
\end{table}

\begin{figure}[t]
    \centering
    \includegraphics[width=3.2in]{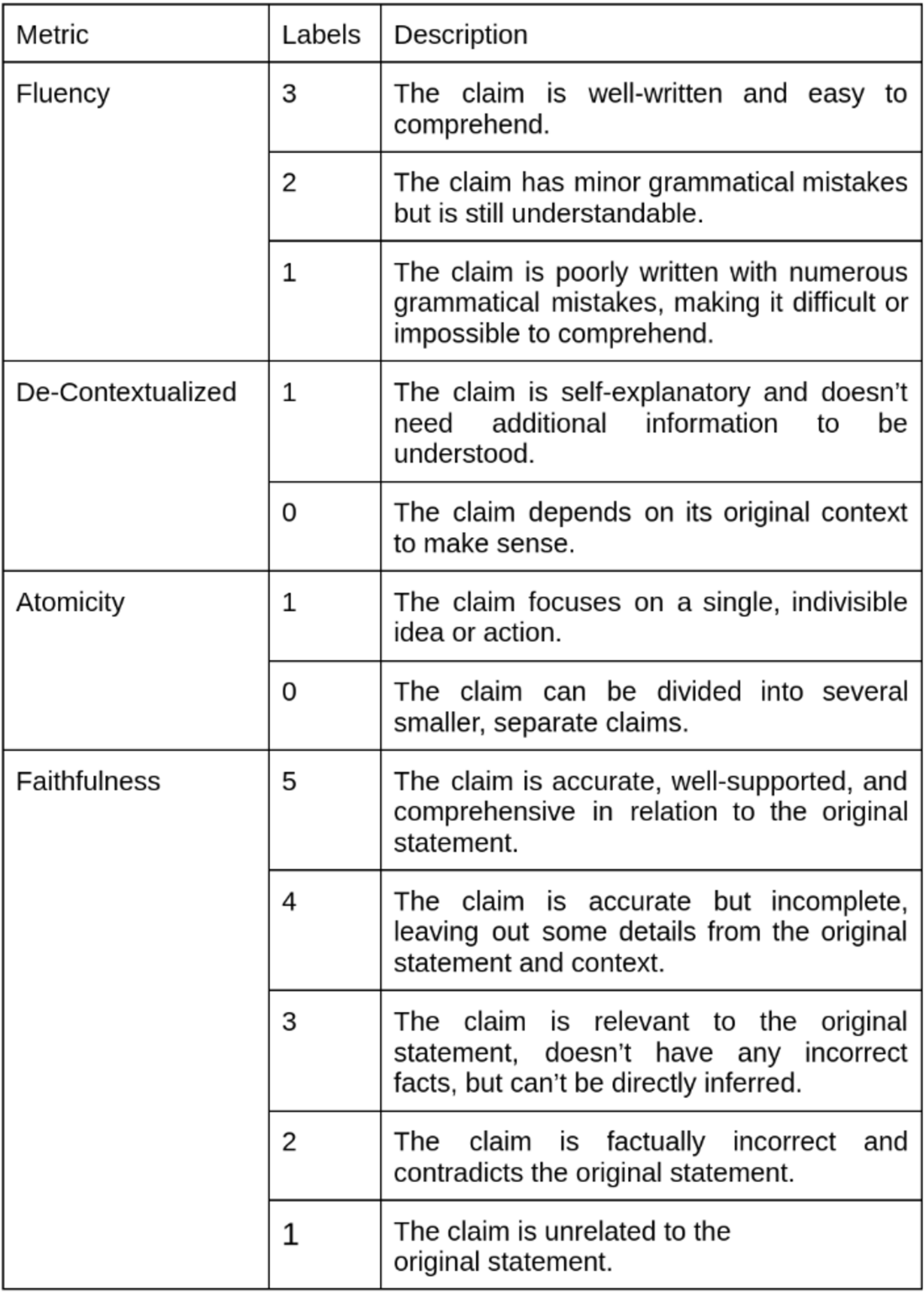}
    \caption{Evaluation Criteria For Claims}
\label{fig:Evaluation Criteria For Claims}
\end{figure}

\subsection{Quality Assessment of Claims and Datasets}
\label{appen_human_eval}
We perform a human annotation to assess the quality of positive and negative claims generated in subsections~\ref{subsec:pos} and \ref{subsec:neg} for our proposed dataset \textit{SciClaimHunt} and \textit{SciClaimHunt\_Num}. This study created a subset of the dataset by randomly selecting one hundred claims generated by claim positive and negative claim generation methods discussed in subsection~\ref{subsec:pos} and \ref{subsec:neg} and providing them to four independent annotators.  We evaluate each claim using the claim evaluation metrics \textit{Fluency}, \textit{De-Contextualization}, \textit{Atomicity}, and \textit{Faithfulness}, as outlined in the study \cite{wright2022generating}. Accordingly, based on these evaluation metrics, each annotator was asked to assign scores to the claims. Figure~\ref{fig:Evaluation Criteria For Claims} presents our annotation instructions, questions to annotators, evaluation criteria for assessing claims, and the scoring for different parameters. To ensure high-quality annotations, we selected annotators with a computer science research background, including research fellows and PhD students. Next, we measure inter-annotator agreement over the score assigned by annotators to one hundred claims for evaluation metrics \textit{Fluency}, \textit{De-Contextualization}, \textit{Atomicity}, and \textit{Faithfulness} of claims using Krippendorff $\boldsymbol{\alpha}$\cite{krippendorff2011computing} and  Fleiss kappa\cite{fleiss1971measuring}. Table~\ref{tab:annot_score} presents inter-annotator agreement score  $\boldsymbol{\alpha}$\cite{krippendorff2011computing} and  Fleiss kappa\cite{fleiss1971measuring} for evaluation metrics \textit{Fluency}, \textit{De-Contextualization}, \textit{Atomicity}, and \textit{Faithfulness}. The Krippendorff and Fleiss kappa scores in Table~\ref{tab:annot_score} indicate substantial agreement among annotators across all metrics, with particularly strong agreement observed for \textit{Atomicity} and \textit{De-contextualization}. Such high agreement for \textit{Atomicity} and \textit{De-contextualization} suggests that the generated claims are generally well-structured, self-contained, and contextually accurate, indicating clarity and completeness of claims. Table~\ref{tab:proposed_dataset} presents the characteristics of our proposed \textit{SciClaimHunt} and  \textit{SciClaimHunt\_Num} datasets.


\begin{figure*}[t]
    \centering
    \includegraphics[width=5in]{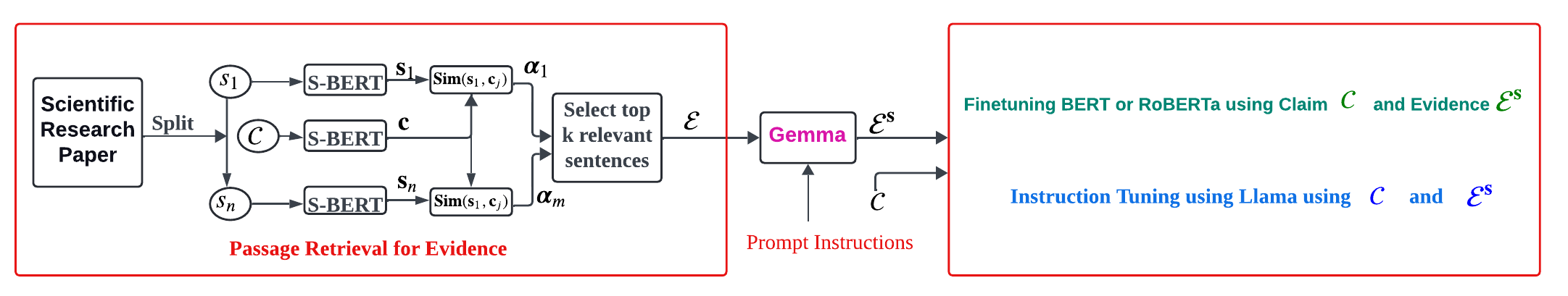}
    \caption{presents a working diagram of the proposed Retrieval-Augmented Generation-based approach for scientific claim validation. Given a research paper $\mathcal{R}$ and a claim $\mathcal{C}$, the research paper $\mathcal{R}$ is first split into a set of passage $\mathcal{S}_i$. Next, the similarity $\boldsymbol{\alpha}_i$ between the encoded representations $\mathbf{s}_i$ and $\mathbf{c}$ of passage $\mathcal{S}_i$ and the claim $\mathcal{C}$, respectively, is computed. Evidence $\mathcal{E}$ is then obtained by selecting the top k passage $\mathcal{S}_i$ with the highest similarity scores $\boldsymbol{\alpha}_i$. This evidence $\mathcal{E}$ along with prompt instruction is passed to the \textit{Gemma} model to generate a fact $\mathcal{E}^s$. Given the claim $\mathcal{C}$ and the generated fact $\mathcal{E}^s$, this study adopts two different approaches for scientific claim validations: (i) fine-tuning BERT and RoBERTa, and (ii) instruction tuning with Llama.}
\label{fig:Rag_model}
\end{figure*}

\section{Proposed Baseline Methods}
Given a research paper $\mathcal{R}$ and claim $C$ the task is to classify each research paper-claim pair $(\mathcal{R}, \mathcal{C})$ into $Y \in \{T, F\}$, where $T$ indicates that claim $\mathcal{C}_j$ is supported by the research paper $\mathcal{R}$, and $F$ indicates that claim $\mathcal{C}$ is refuted by the research paper $\mathcal{R}$.


\subsection{Evidence Retrieval}
\label{subsec:evi_ret_val}
Research papers are structured into sections containing numerous sentences and paragraphs, but transformer-based and large language models (LLMs) have limited context lengths. Therefore, extracting the passage from the research paper $\mathcal{R}$ that directly discusses the claim $\mathcal{C}$ is essential. With this motivation, given a pair of research papers and claim $(\mathcal{R}, \mathcal{C})$, we first extract the relevant passage as evidence $\mathcal{E}$ from the research paper $\mathcal{R}$ that are contextually similar to $\mathcal{C}$ and contain direct discussions related to $\mathcal{C}$. We first split the research paper $\mathcal{R}$ into a set of passage $\mathcal{R}^+ = {\mathcal{S}_1, \mathcal{S}_2, ..., \mathcal{S}_n}$ and obtain encoded representations $\mathbf{s}_i$ for the $i^{th}$ sentence in the set $\mathcal{R}^+$ and $\mathbf{c}$ for the claim $\mathcal{C}$ using Sentence-BERT (SBERT)\cite{reimers-2019-sentence-bert}. Next, we estimate the cosine similarity $\boldsymbol{\alpha}_i$ between the encoded representations $\mathbf{s}_i$ and $\mathbf{c}$.  Subsequently, we form a vector $\mathbf{a}$, where the $i^{th}$ element in the vector $\mathbf{a}$ represents the similarity $\boldsymbol{\alpha}_i$ between the passage $\mathbf{s}_i$ and the claim $\mathbf{c}$. We then sort the vector $\mathbf{a}$ in decreasing order and select the top $k$ passage $\mathcal{S}_i$ corresponding to the highest values of $\boldsymbol{\alpha}_i$ from the research paper $\mathcal{R}$ as evidence $\mathcal{E}$ which support or refutes claim $\mathcal{C}_j$.   

\subsection{Retrieval-Augmented Generation (RAG) Approach}
Given the extensive text within the various sections of a research paper and the token limitations of transformer models such as BERT and RoBERTa, utilizing an entire paper for finetuning transformer-based models for scientific claim validation tasks is impractical. Motivated by the above limitation, this study adopts a Retrieval-Augmented Generation (RAG) approach for scientific claim validations. Given a claim and research paper pair $(\mathcal{C},\mathcal{R})$, our Retrieval-Augmented Generation (RAG)~\cite{lewis2020retrieval} based approach for scientific claim validation first extracts evidence $\mathcal{E}$ from the research paper $\mathcal{R}$, using the evidence retrieval method described in subsection~\ref{subsec:evi_ret_val}. Next, the extracted evidence $\mathcal{E}$ along with prompt instructions, is provided as context to a \textit{Gemma} model~\cite{team2024gemma}\footnote{\href{https://huggingface.co/google/gemma-2-2b-it}{Google/gemma-2-2b-it}}~, which generates a summary $\mathcal{E}^s$ by rephrasing, summarizing, and presenting it as factual statements. The key motivation for applying LLMs to the extracted evidence $\mathcal{E}$ to generate evidence summary $\mathcal{E}^s$ is that the passage in $\mathcal{E}$ is sourced from various sections of the research paper based on their similarity to the claim. This results in a collection of independent passages rather than a coherent or sequential narrative. Therefore, the extracted evidence $\mathcal{E}$ is provided as context to the \textit{Gemma} model, which rephrases and generates a summary as factual statements in a coherent sequence of sentences, $\mathcal{E}^s$, which is facts and summary generated using passages in $\mathcal{E}$. Next, given the pair $(\mathcal{C}, \mathcal{E}^s)$, we fine-tune  BERT and RoBERTa and also propose instruction tuning with a large language model  Llama for claim verification.\newline 
\textbf{Scientific Claim Verification using BERT and RoBERTa}: First, we obtain the encoded representations $\mathbf{c}$ and $\mathbf{e}^s$ for claim $\mathcal{C}$  and summary of the retrieved evidence $\mathcal{E}^s$ using BERT or RoBERTa models, respectively. Next, estimate the similarity and dissimilarity feature vector between the encoded representations $\mathbf{c}$ and $\mathbf{e}^s$ using the steps outlined in subsection~\ref{subsec:feature}, and pass this feature vector to a two-layer neural network for scientific claim verification.\newline
\textbf{Instruction Tuning with Llama :}
Given a pair consisting of the claim $\mathcal{C}$ and the summary of extracted evidence $\mathcal{E}^s$, along with a prompt instruction $\mathcal{P}_r$, we fine-tune the Llama~\cite{touvron2023llama}\footnote{\href{https://huggingface.co/meta-llama/Llama-3.2-3B-Instruct}{Meta-llama/Llama-3.2-3B-Instruct}} model for scientific claim verification. Our prompt instruction $\mathcal{P}_r$ is as follows: $\mathcal{P}_r$ = \textit{You are an expert in making judgments. You are given a claim and some sentences. Your job is to verify whether the evidence encapsulated in retrieved sentences supports the claim. The retrieved sentences come from a research paper. If the evidence encapsulated in the retrieved sentences supports the claim, you need to answer as 'positive' otherwise 'negative'}.
\begin{figure}[t]
    \centering
    \includegraphics[width=3in]{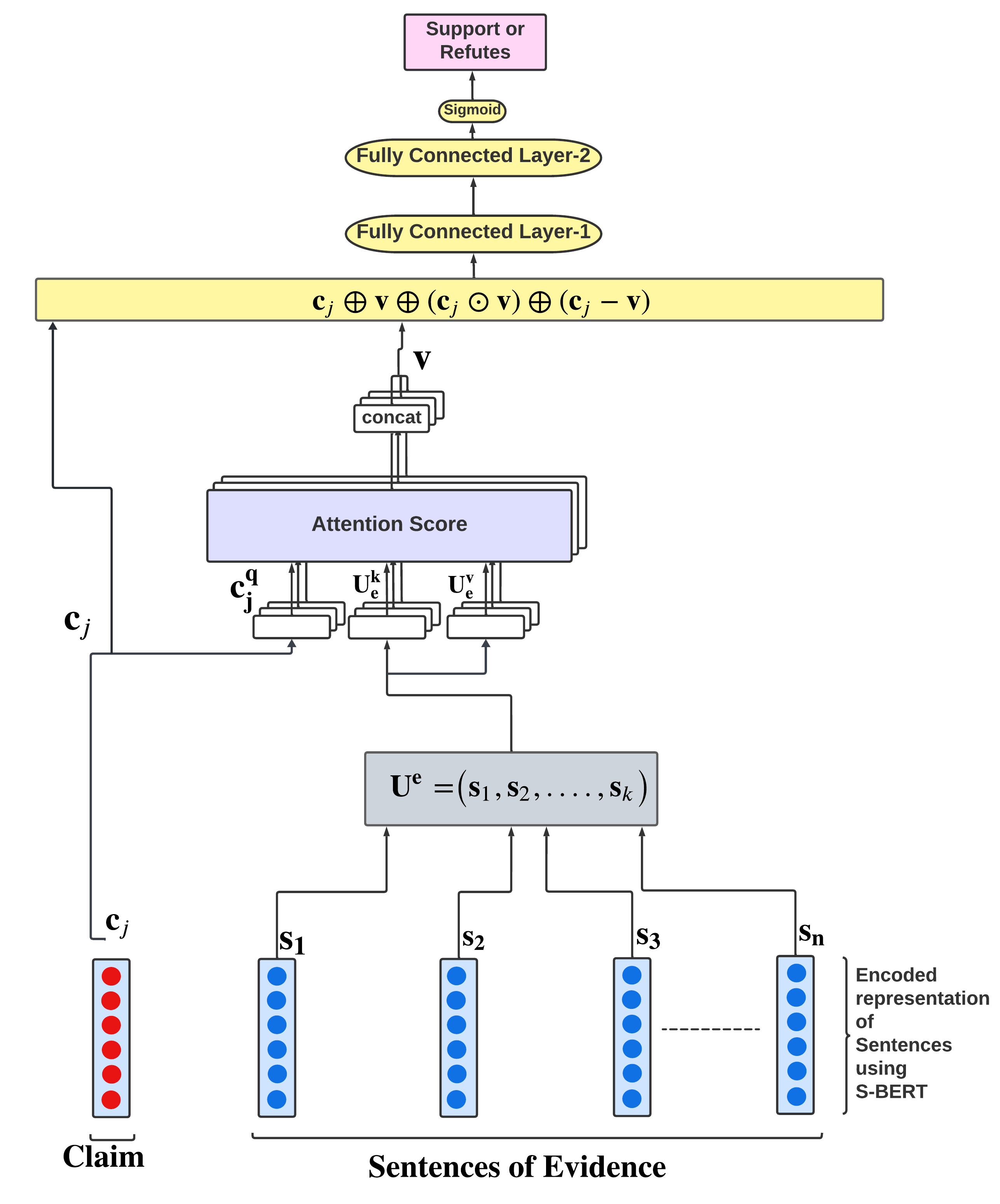}
    \caption{present the \textbf{Claim-Evidence Matching using Multi-Head Attention (CEM)} for scientific claim verification. First, the claim $\mathcal{C}_j$ and sentences $\mathcal{S}_i$ are encoded using \textit{S-BERT} to obtain the encoded representations $\mathbf{c}_j$ for the claim $\mathcal{C}_j$ and $\mathbf{s}_i$ for the $i^{th}$ sentence in the evidence set. Next, we construct an evidence representation matrix $\mathbf{U}^e$ by stacking the encoded representations $\mathbf{s}_i$ of all sentences in the evidence set, where each row of $\mathbf{U}^e$ corresponds to the encoded representation of an individual sentence. Subsequently, we apply multi-head attention between the claim and the evidence set, using the encoded representation of the claim $\mathbf{c}_j$ as the query and the evidence matrix $\mathbf{U}^e$ as both the key and value, to obtain the evidence representation vector $\mathbf{v}$ based on the similarity between the claim and the evidence. Finally, we estimate the similarity and difference feature vector between $\mathbf{C}_j$ and $\mathbf{v}$, which is passed through two fully connected layers for classification.} 
    \label{fig:cem_model}
\end{figure}
\subsection{Claim-Evidence Matching using Multi-Head Attention (CEM)}
\label{subsec:CEM}
 A scientific report or paper is structured into sections, each serving a specific purpose. When evaluating claims, extracting evidence from each relevant section is crucial, ensuring that context-specific information supports or refutes the claims. This study proposes \textbf{CEM} method, which improves upon traditional evidence extraction and retrieval methods discussed in the subsection~\ref{subsec:evi_ret_val} by retrieving evidence from each section rather than merely selecting passages with the highest similarity to the claim, regardless of the section. This section-wise retrieval helps to collect more effective evidence from each section, which helps the claim verification model to make a more informed and accurate decision. Figure~\ref{fig:cem_model} illustrates our proposed Claim-Evidence Matching using Multi-Head Attention (CEM) method for scientific claim verification. The CEM method first divides the research paper $\mathcal{R}$ into a set of sections $\mathcal{U}^+ = {\mathcal{U}_1, \mathcal{U}_2, ..., \mathcal{U}_t}$, where $t$ represents the total number of sections in the paper $\mathcal{R}$. Given a pair consisting of a claim $\mathcal{C}_j$ and a section $\mathcal{U}_i$ from the research paper $\mathcal{R}$, we extract passage highly similar to the claim as evidence $\mathcal{U}_i^e$ from each section $\mathcal{U}_i$ using the evidence retrieval methods outlined in subsection~\ref{subsec:evi_ret_val}. Finally, the extracted evidence from each section is merged to form the evidence $\mathcal{U}^e = \mathcal{U}_1^e \oplus \mathcal{U}_2^e \oplus ... \oplus \mathcal{U}_t^e$.  Additionally, the evidence set $\mathcal{U}^e$ is divided into individual sentences $\mathcal{S}_1, \mathcal{S}_2, ..., \mathcal{S}_k$, and we obtain an encoded representation $\mathbf{s}_i$ for each sentence $\mathcal{S}_i \in \mathcal{U}^e $ from the evidence set $\mathcal{U}^e$ for all $i=1$ to $k$, along with an encoded representation $\mathbf{c}$ for the claim $\mathcal{C}$ using Sentence-BERT (SBERT)\cite{reimers-2019-sentence-bert}. We apply multi-head attention\cite{vaswani2017attention} between the claim and the evidence set, using the encoded representation of the claim $\mathbf{c}$ as the query and the evidence matrix $\mathbf{U}^e$ as both the key and value, to obtain the evidence representation vector $\mathbf{v}$ based on the similarity between the claim and the evidence. The prime motivation behind applying multi-head attention between claim and evidence are as follows: (i) If the evidence supports the claim, the multi-head attention will assign high weights to sentences that are highly similar to the claim, emphasizing their importance in generating the representative vector $\mathbf{v}$. (ii) Similarly, If the evidence extracted from the research paper refutes the claim, the sentences in the evidence set will either be unrelated or contradictory to the claim, leading the multi-head attention to assign low weights, indicating their lack of importance in generating the representative vector $\mathbf{v}$. Subsequently, We compute the similarity and dissimilarity feature vector $\mathbf{f}$ between the claim $\mathbf{c}_j$ and evidence $\mathbf{v}$ (see subsection~\ref{subsec:feature}) and pass it through a two-layer neural network for scientific claim validations.


\subsubsection{Graph-based Claim and Evidence Matching~(\textbf{GCEM}) }
Representing a research paper as a graph captures relationships and structure between sentences, sections, and the entire paper, enhancing contextual understanding and revealing both local and global patterns\cite{ nikolentzos2020message, zhang2020every}. Additionally, graph representations effectively manage non-sequential and complex interactions, improving analysis of the paper's structure and content\cite{ nikolentzos2020message, zhang2020every}. Motivated by the advantages of representing documents as graphs, we propose \textit{Graph-based Claim and Evidence Matching} (\textbf{GCEM}) for scientific claim verification. The key advantage of GCEM is that it considers the entire research paper $\mathcal{R}$ as evidence to support or refute the claim $\mathcal{C}$ rather than relying on a few selected passages as evidence. Next, we construct a research paper graph $\mathcal{R}_g = \{\mathcal{V}, \mathcal{E}\}$ by considering each sentence $\mathcal{S}_i$ as a node and assigning the encoded representation $\mathbf{s}_i$ as the initial node embedding.  We then estimate the cosine similarity $x_{ij}$ between two nodes $\mathcal{S}_i$ and $\mathcal{S}_j$ in the graph $\mathcal{R}_g$ using their initial node embeddings $\mathbf{s}_i$ and $\mathbf{s}_j$. Next, we add an edge between two nodes $\mathcal{S}_i$ and $\mathcal{S}_j$ in the research paper graph $\mathcal{R}_g$ if $x_{ij} \geq \boldsymbol{\beta}$, where $\beta$ is a user-defined similarity threshold. Subsequently, we form a node embedding matrix $\mathbf{B}$, which represents the node embeddings of the research paper graph $\mathcal{R}_g$, where the $i^{th}$ row corresponds to the initial node embedding $\mathbf{s}_i$ of node $\mathcal{S}_i$ in the graph $\mathcal{R}_g$. Given the node embedding matrix $\mathbf{B}$ of the research paper graph $\mathcal{R}_g$, we apply a $\mathbf{t}$-layer Graph Attention Network~(GAT)~\cite{velickovic2018graph} to the research paper graph $\mathcal{R}_g$ and obtain the transformed node embedding matrix $\mathbf{B}^\mathbf{l}$ for the research paper $\mathcal{R}_g$. The primary motivation for applying GAT over $\mathcal{R}_g$ is to update the node embedding of each node in $\mathcal{R}_g$ based on its local neighborhood structure and the context of its neighboring nodes. Our GAT implementation follows the settings of GAT outlined in study~\cite{velickovic2018graph}. Next, we obtain the research paper graph representation feature vector $\mathbf{r}$ by concatenating the vectors obtained after applying min ({\it Min}), max ({\it Max}), and average ({\it AVG}) pooling operations over the transformed node embedding matrix $\mathbf{B}^{l}$. Subsequently, we estimate the similarity and dissimilarity feature vector $\mathbf{f}$ between the encoded representation $\mathbf{c}$ and research paper graph representation vector $\mathbf{r}$ of the research graph $\mathcal{R}_g$ and pass it to a two-layer neural network for classification using the steps outlined in subsection~\ref{subsec:feature}



\subsection{Similarity and Dissimilarity Feature Estimation and Classification}
\label{subsec:feature}
This study estimates the similarity and dissimilarity features between the encoded representations $\mathbf{c}$ and $\mathbf{e}_i^s$ by calculating the angle $\mathbf{f}^{\boldsymbol+}$ and the difference $\mathbf{f}^{\boldsymbol-}$ between $\mathbf{c}$ and $\mathbf{e}_i^s$, as defined by the following equations:
\begin{equation} \label{eq:claim_body_evidencee_diff}
\mathbf{f}^{\boldsymbol+}\; \boldsymbol{,} \;\mathbf{f}^{\boldsymbol-}  \;  \text{= }\;   \mathbf{e}_i^s  \boldsymbol{\odot}  \mathbf{c}_j\; \,  \; \mathbf{e}_i^s  \boldsymbol{-} \mathbf{c}_j
\end{equation}
Here, $\boldsymbol{\odot}$ represents element-wise multiplication between two vectors, and $ \boldsymbol{-}$ represents element-wise difference between two vectors. Subsequently, we form a feature vector $\mathbf{f}$ using Equation~\ref{final_feature_vec} as defined below.

\begin{equation} \label{final_feature_vec}
\mathbf{f} \text{ = } \big(\mathbf{f}^{\boldsymbol+} \boldsymbol{\oplus}  \mathbf{f}^{\boldsymbol-} \boldsymbol{\oplus}\mathbf{e}_i^s \boldsymbol{\oplus} \mathbf{c}_j  \big)
\end{equation}
Where $\boldsymbol{\oplus}$ represents the vector concatenation operator. Once we obtain the feature vector $\mathbf{f}$, we pass it through a two-layer neural network, and we use the cross-entropy loss function to learn the model parameters.



\section{Experimental Results and Discussions}

\begin{table}[t]
\centering
\caption{Details of hyperparameter settings used to generate the results.}
\label{tab:hyper_values}
\renewcommand{\arraystretch}{1.2} 
\setlength{\tabcolsep}{6pt} 
\begin{adjustbox}{max width=\columnwidth}
\begin{tabular}{l l} 
\hline
\textbf{Hyperparameters}        & \textbf{Values}            \\  
\hline
Epoch                           & 100                        \\  
Threshold value $\boldsymbol{\beta}$ & 0.25, 0.5, 0.75         \\  
No. of Attention Heads          & 8                          \\  
Batch Size                      & 2, 64                      \\  
Embedding Dimension             & 384                        \\  
Learning Rate                   & 0.01                       \\  
Loss Function                   & Cross Entropy              \\  
No. of Layers in GAT            & 3                          \\  
 No. of sentences in a research paper $\mathbf{R}$            & 265                        \\ 
No. of Sections in a research paper $\mathbf{R}$            & 5                \\ 
\hline
\end{tabular}
\end{adjustbox}
\vspace{-2mm} 
\end{table}

\begin{table}[t]
\centering
\caption{Characteristics of \textbf{SCIFACT} and \textbf{SCIFACT-OPEN} datasets. 
Here, \#Claim and \#Abstract indicate the average number of words in the Claim and the abstract of a scientific research paper, respectively. Similarly, \#Sent indicates the average number of sentences in a scientific research paper.}
\label{tab:test_dataset}
\renewcommand{\arraystretch}{1.2} 
\setlength{\tabcolsep}{5pt} 
\begin{adjustbox}{max width=\columnwidth}
\begin{tabular}{l r r r r r r} 
\hline
\textbf{Dataset}       & \textbf{Support} & \textbf{Refutes} & \textbf{Total} & \textbf{\#Claims} & \textbf{\#Abstract} & \textbf{\#Sent} \\  
\hline
\textbf{SCIFACT-OPEN}  & 122              & 112              & 234            & 11.4              & 49.1                & 3.82            \\  
\textbf{SCIFACT}       & 830              & 463              & 1,293          & 12.19             & 30.12               & 2.11            \\  
\hline
\end{tabular}
\end{adjustbox}
\vspace{-2mm} 
\end{table}

\subsection{Experimental Setups}
This study considers Accuracy (Acc.), F-measure (F.), and class-wise F-measure as performance metrics to study the performance of proposed and baseline models in scientific claim verification. Table~\ref{tab:hyper_values}  presents the details of the experimental hyper-parameters used to generate the results presented in this paper. This study also considers the scientific claim verification datasets \textit{SCIFACT}\cite{wadden2020fact} and  \textit{SCIFACT-OPEN}\cite{wadden2022scifact} from literature to evaluate the performance of the proposed baseline models. Table~\ref{tab:test_dataset} presents the characteristics of \textit{SCIFACT} and  \textit{SCIFACT-OPEN}.  

\begin{table}[t]
\centering
\caption{Performance of the proposed baseline models on the dataset. 
Here, \textbf{(Acc.)}, \textbf{(S.)}, and \textbf{(R.)} indicate the \textbf{accuracy} and F-measure for the \textbf{Support} and \textbf{Refutes} classes, respectively. 
\textbf{ER} refers to the \textbf{Evidence Retrieval} approach, \textbf{RAG} stands for the \textbf{Retrieval Augmented Generation} approach, and \textbf{FRP} indicates the approach where models consider the \textbf{Full Research Paper} as evidence.}
\label{tab:main}
\renewcommand{\arraystretch}{1.3} 
\setlength{\tabcolsep}{4pt} 
\begin{adjustbox}{max width=\columnwidth}
\begin{tabular}{l l c c c} 
\hline
\textbf{Approach}    & \textbf{Model}            & \textbf{Acc.} & \textbf{S.} & \textbf{R.} \\  
\hline
\multirow{3}{*}{ER}  & BERT                     & \textcolor{magenta}{\textbf{0.985}} & \textcolor{magenta}{\textbf{0.982}} & \textbf{0.986} \\  
                              & RoBERTa                  & 0.934 & 0.935 & 0.931 \\  
                              & CEM                      & 0.904 & 0.889 & 0.915 \\  
\hline
\multirow{3}{*}{RAG} & BERT                     & 0.982 & 0.979 & 0.984 \\  
                              & RoBERTa                  & 0.929 & 0.934 & 0.923 \\  
                              & Llama                    & \textbf{0.983} & \textbf{0.980} & \textcolor{magenta}{\textbf{0.990}} \\  
\hline
\multirow{4}{*}{FRP} & CEM                      & \textbf{0.935} & \textbf{0.923} & \textbf{0.944} \\  
                              & GCEM($\boldsymbol\beta \; \text{=}\; 0.25$) & 0.840 & 0.860 & 0.820 \\  
                              & GCEM($\boldsymbol\beta \; \text{=}\; 0.5$)  & 0.892 & 0.892 & 0.891 \\  
                              & GCEM($\boldsymbol\beta \; \text{=}\; 0.75$) & \text{0.921} & \textbf{0.923} & \text{0.922} \\  
\hline
\end{tabular}
\end{adjustbox}
\vspace{-2mm} 
\end{table}

\subsection{Results and Discussions}
Table~\ref{tab:main} presents the performance of the proposed baseline model on our dataset. In the Table, proposed baseline models are grouped into three Approaches: ER, RAG, and FRP. (i) Evidence Retrieval (\textbf{ER}): In this approach, the model only considers the passages extracted from the research paper using a dense retrieval method (discussed in subsection~\ref{subsec:evi_ret_val}) as evidence. (i) Evidence Retrieval (\textbf{ER}): In this approach, the model only considers the passages extracted from the research paper using a dense retrieval method (discussed in subsection~\ref{subsec:evi_ret_val}) as evidence. (ii) Retrieval Augmented Generation (\textbf{RAG}): The model retrieves passage relevant to the claims from the research paper (discussed in subsection~\ref{subsec:evi_ret_val}) and passes them through a \textit{Gemma} as context to generate a summary and facts, generated summary and fact is then considered evidence. (iii) Full Research Paper (\textbf{FRP}): In this approach, the models consider the entire research paper, excluding the conclusion section, as evidence. From Table~\ref{tab:main}, it is apparent that the performance of the \textbf{ER} method is superior with the BERT model compared to the performance of \textbf{ER} methods with CEM and Roberta and the performance of \textbf{RAG}-based methods is superior with  Llama model. Similarly, the performance of \textbf{RAG}-based methods is superior with the Llama model compared to that of \textbf{RAG} with Roberta and BERT.  For the Full Research Paper (\textbf{FRP}) approaches, GCEM outperforms the CEM model. Additionally, GCEM with $\beta = 0.75$ shows better results than $\beta = 0.25$ and $\beta = 0.5$, indicating that $\beta = 0.75$ effectively captures relationships between sentence nodes in the research paper. For the Full Research Paper (\textbf{FRP}) approaches, GCEM outperforms the CEM model. Furthermore, GCEM with $\beta = 0.75$ outperforms configurations with  $\beta = 0.25$ and $\beta = 0.5$. This suggests that $\beta = 0.75$ effectively captures relationships between sentence nodes because a higher $\beta$ value results in a sparser graph, connecting only sentences with strong contextual similarity. The performance comparison of \textbf{ER}, \textbf{RAG}, and \textbf{FRP} reveals the following observations:  ER and RAG, which leverage evidence retrieval and generation, respectively, achieve superior performance compared to FRP. From such observation, we conclude that methods focused on identifying and extracting relevant evidence from a large scientific document or entire research paper, whether through retrieval or generation, are more effective for scientific claim verification than approaches that consider the whole research paper or scientific document as evidence as evidence.

\begin{table}[t]
\centering
\caption{Performance of the proposed baseline models, trained on the proposed dataset and tested on \textbf{SCIFACT} \cite{wadden2020fact} and \textbf{SCIFACT-OPEN} \cite{wadden2022scifact}.}
\label{tab:val_data_res}
\renewcommand{\arraystretch}{1.3} 
\setlength{\tabcolsep}{4pt} 
\begin{adjustbox}{max width=\columnwidth}
\begin{tabular}{l l c c c c c c} 
\hline
\multirow{2}{*}{\textbf{Approach}} & \multirow{2}{*}{\textbf{Model}} & \multicolumn{3}{c}{\textbf{SCIFACT}} & \multicolumn{3}{c}{\textbf{SCIFACT-OPEN}} \\ 
\cline{3-8} 
                                   &                                 & \textbf{Acc.} & \textbf{T}    & \textbf{F.}   & \textbf{Acc.}  & \textbf{T}     & \textbf{F.}    \\  
\hline
\multirow{3}{*}{ER}       & BERT                  & \textbf{0.728} & \textbf{0.802} & \textbf{0.568} & 0.695          & 0.714          & 0.634          \\  
                                   & RoBERTa               & \textbf{0.728} & \textbf{0.802} & \textbf{0.568} & \textbf{0.717} & \textbf{0.760} & \textbf{0.656} \\  
                                   & CEM                   & 0.635          & 0.769          & 0.132          & 0.521          & 0.670          & 0.125          \\  
\hline
\multirow{3}{*}{RAG}      & BERT                  & 0.696          & 0.795          & 0.413          & 0.615          & 0.701          & 0.457          \\  
                                   & RoBERTa               & 0.713          & 0.765          & 0.656          & 0.707          & 0.743          & 0.653          \\  
                                   & Llama                 & \textbf{0.792} & \textbf{0.837} & \textbf{0.711} & \textbf{0.748} & \textbf{0.754} & \textbf{0.741} \\  
\hline
\multirow{4}{*}{FRP}      & CEM                   & 0.652          & 0.780          & 0.163          & 0.517          & 0.658          & 0.175          \\  
                                   & $\mathbf{GCEM} (\boldsymbol{\beta} \text{=}0.25)$ & 0.754 & 0.754 & 0.746 & 0.743          & 0.748          & 0.739          \\  
                                   & $\mathbf{GCEM} (\boldsymbol{\beta} \text{=}0.5)$  & 0.782 & 0.789 & 0.745 & 0.769          & 0.823          & 0.749          \\  
                                   & $\mathbf{GCEM} (\boldsymbol{\beta} \text{=}0.75)$ & \textcolor{magenta}{\textbf{0.823}} & \textcolor{magenta}{\textbf{0.847}} & \textcolor{magenta}{\textbf{0.841}} & \textcolor{magenta}{\textbf{0.787}} & \textcolor{magenta}{\textbf{0.798}} & \textcolor{magenta}{\textbf{0.778}} \\  
\hline
\end{tabular}
\end{adjustbox}
\vspace{-2mm} 
\end{table}

\begin{table}[t]
\centering
\caption{Statistical comparison between the \textbf{SCIFACT} and \textbf{SCIFACT-OPEN} datasets in terms of the number of sentences in the evidence. \textbf{STD} refers to the Standard Deviation.}
\label{tab:sent_dstributions}
\begin{adjustbox}{width=\columnwidth}
\begin{tabular}{l l ll} 
\hline
\textbf{Corpus}  & \textbf{SCIFACT} & \textbf{SCIFACT-OPEN}     & \textbf{Proposed Dataset} \\ 
\hline
\textbf{MEAN}    & 1.41             & 1.76                     & 105.98                    \\ 
\hline
\textbf{STD}     & 0.65             & 0.56                     & 33.58                     \\ 
\hline
\textbf{Minimum} & 1                & 1                        & 24                        \\ 
\hline
\textbf{Maximum} & 5                & 7                        & 300                       \\ 
\hline
\textbf{25\%}    & 1                & 1                        & 83.3                      \\ 
\hline
\textbf{50\%}    & 1                & 2                        & 103                       \\ 
\hline
\textbf{75\%}    & 2                & 3                        & 140                       \\ 
\hline
\end{tabular}
\end{adjustbox}
\end{table}

\subsubsection{Validation of Proposed Datasets}
To evaluate both the real-world applicability and generalizability of our proposed datasets, we assessed the performance of our baseline model trained using proposed datasets over existing scientific claim verification datasets from the literature. With this motivation, we trained the proposed baseline models using \textit{SciClaimHunt} and evaluated their performance by using \textit{SCIFACT}\cite{wadden2020fact} and \textit{SCIFACT-OPEN}\cite{wadden2022scifact} as test datasets. Table~\ref{tab:val_data_res} presents the performance of the proposed baseline models trained on the \textit{SciClaimHunt} datasets and evaluated using \textit{SCIFACT}\cite{wadden2020fact} and \textit{SCIFACT-OPEN}\cite{wadden2022scifact} as test datasets. From Table~\ref{tab:val_data_res}, it is evident that the  \textbf{RAG} approach with Llama and $GCEM (\beta = 0.75)$ demonstrates outstanding performance when trained on the \textit{SciClaimHunt} dataset and evaluated using \textit{SCIFACT} and \textit{SCIFACT-OPEN} as test datasets.  From such observation from Table~\ref{tab:val_data_res}, we can conclude that the models trained on our proposed dataset (\textit{SciClaimHunt}) are generalizable on unseen scientific claim verification datasets and effective for scientific claim verification in real-world scenarios. Consequently, we conclude that our proposed dataset (\textit{SciClaimHunt}) is highly high-quality and reliable for training models for scientific claim verification tasks.
  
\begin{figure}[]
  \centering
    \includegraphics[width=0.5\textwidth, height=0.45\textheight]{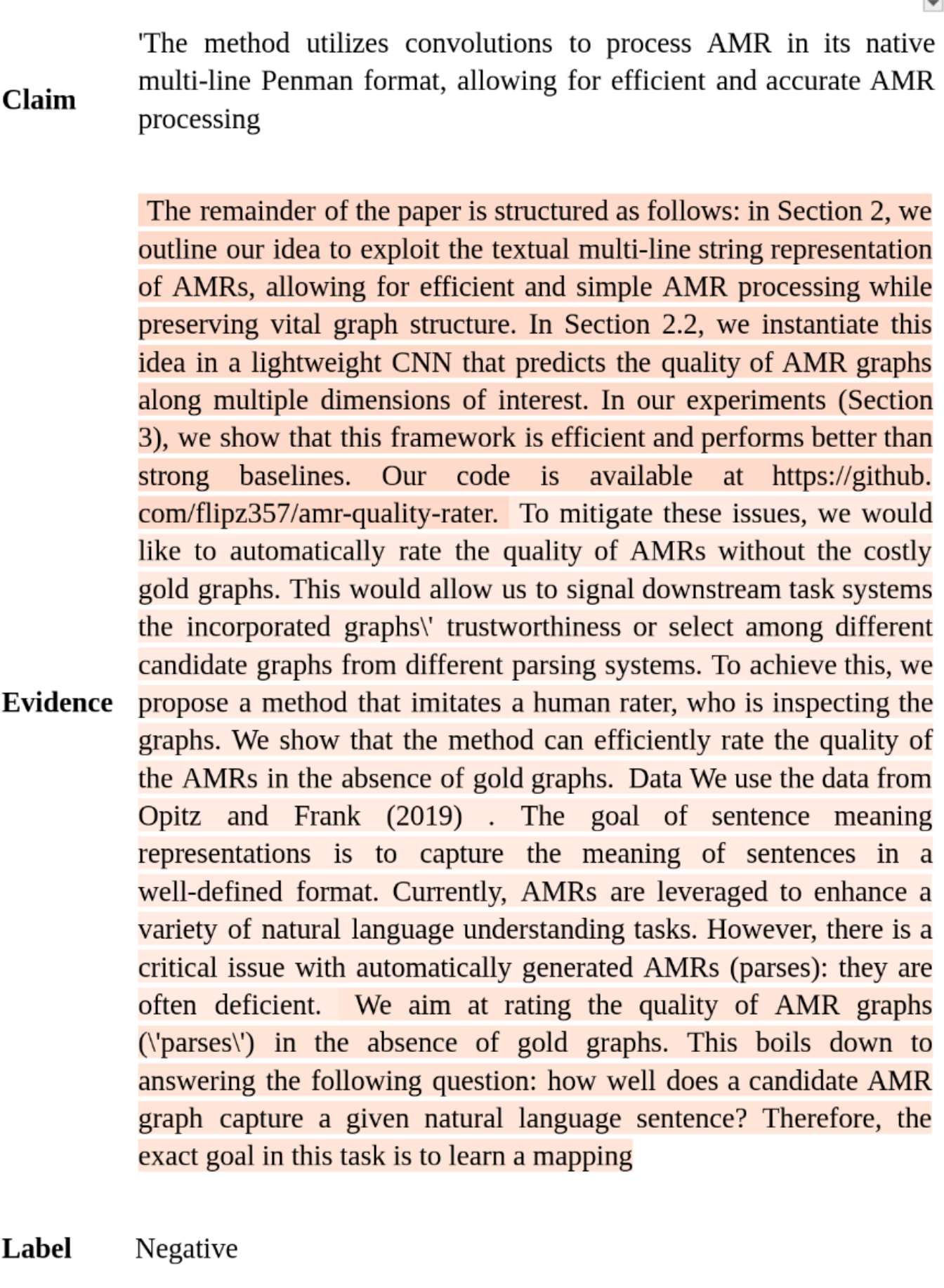}
    \caption{presents attention heatmaps showing the attention weights between the claim and various sentences of the evidence for the \textit{Refutes} class (negative claims) samples. The heatmaps reveal that the multi-head attention component of the CEM model assigns moderate attention weights to sentences that are just related and neural to the claim and lower weights to sentences with minimal relevance to the claim. The darker colour signifies the higher attention weight assigned to the respective sentence from the evidence set and vice-versa. }  
\label{fig:neg_heatmap_2}
\end{figure}

  However, from Table~\ref{tab:val_data_res}, it is apparent that the performance of the CEM model declines when trained on the proposed dataset and tested on \textit{SCIFACT} and \textit{SCIFACT-OPEN}. We further study the reason behind the declines in the performance CEM and found that the number of sentences in the evidence sets of \textit{SCIFACT} and \textit{SCIFACT-OPEN} is significantly lower than the number of sentences in the evidence set of the proposed dataset. Since CEM employs multi-head attention between claim and sentences of the evidence set, the weight matrix in the multi-head attention component of CEM is dependent on the number of sentences in the evidence set, which varies between the proposed dataset and \textit{SCIFACT} and \textit{SCIFACT-OPEN}. The difference in terms of the number of sentences in the proposed dataset \textit{SciClaimHunt} and \textit{SCIFACT} and \textit{SCIFACT-OPEN} could be a possible reason behind the decline in the performance of CEM models over  \textit{SCIFACT} and \textit{SCIFACT-OPEN}. The number of sentences in proposed dataset \textit{SciClaimHunt} and \textit{SCIFACT} and \textit{SCIFACT-OPEN} is different because \textit{SciClaimHunt} consider entire research paper as evidence which supports and refutes the claim whereas \textit{SCIFACT} and \textit{SCIFACT-OPEN} considers only abstract of the research paper as evidence. Table~\ref{tab:val_data_res} shows the sentence distribution differences between the proposed dataset \textit{SciClaimHunt} and \textit{SCIFACT} and \textit{SCIFACT-OPEN}.

\begin{table}[t]
\centering
\caption{Performance of the proposed baseline models on the \textit{SciClaimHunt\_Num} dataset.}
\label{tab:num_res}
\renewcommand{\arraystretch}{1.2} 
\setlength{\tabcolsep}{5pt} 
\begin{adjustbox}{max width=\columnwidth}
\begin{tabular}{l l c} 
\hline
Approach             & Model                            & \textbf{Accuracy}      \\ 
\hline
\multirow{3}{*}{ER}  & BERT                           & \textcolor{magenta}{\textbf{0.985}} \\ 
                     & RoBERTa                        & 0.927                  \\ 
                     & CEM                            & 0.890                  \\ 
\hline
\multirow{3}{*}{RAG} & BERT                           & 0.982                  \\ 
                     & RoBERTa                        & 0.919                  \\ 
                     & Llama                          & \textbf{0.980}         \\ 
\hline
\multirow{4}{*}{FRP} & CEM                            & \textbf{0.948}         \\ 
                     & GCEM($\beta = 0.25$)           & 0.819                  \\ 
                     & GCEM($\beta = 0.5$)            & 0.886                  \\ 
                     & GCEM($\beta = 0.75$)           & \textbf{0.927}         \\ 
\hline
\end{tabular}
\end{adjustbox}
\vspace{-2mm} 
\end{table}

\begin{figure}[]
  \centering
    \includegraphics[width=0.5\textwidth, height=0.41\textheight]{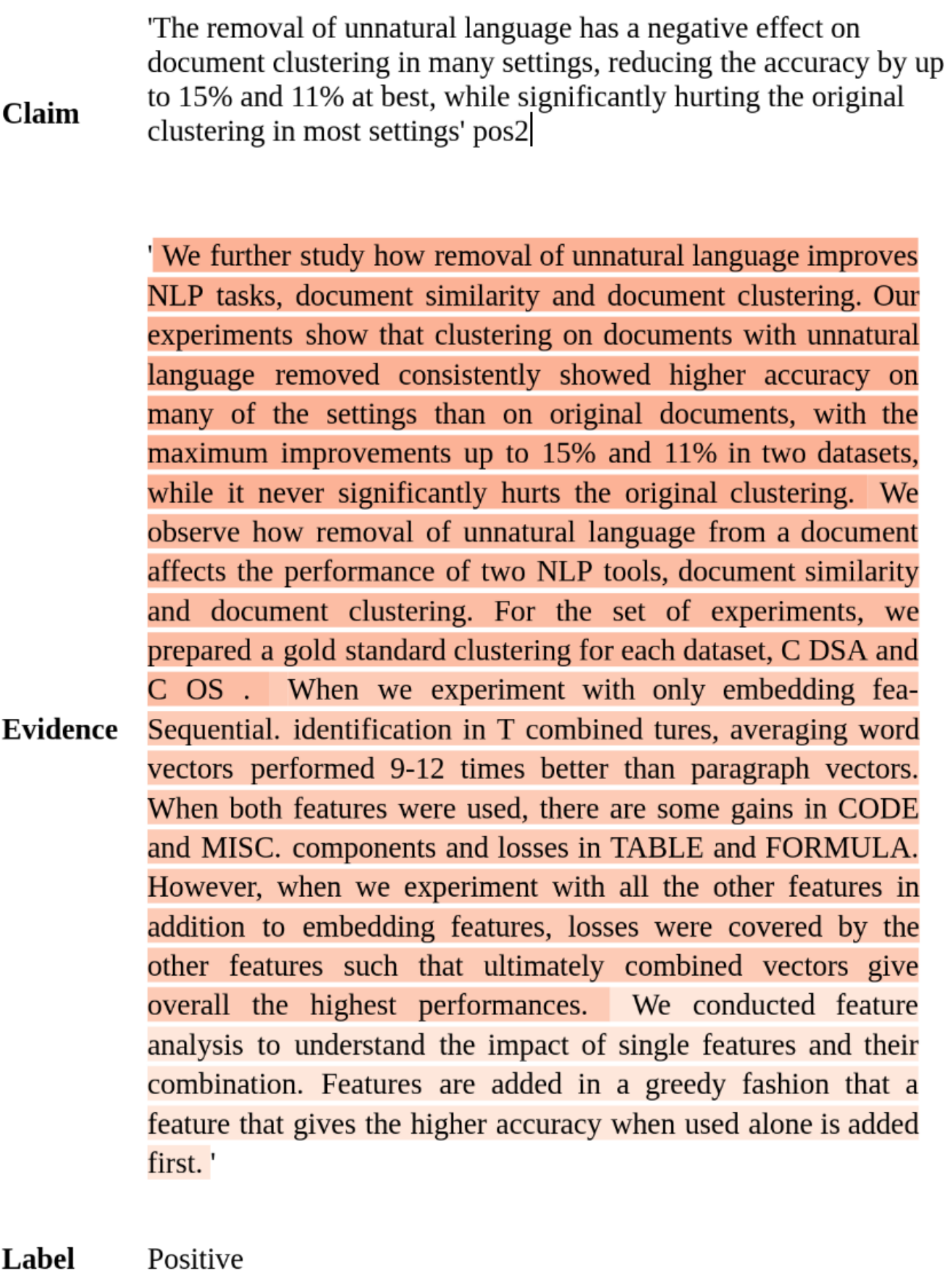}
    \caption{ presents attention heatmaps showing the attention weights between the claim and various sentences of the evidence for the \textit{Supporting} class (positive claims) samples. The heatmaps reveal that the multi-head attention component of the CEM model assigns higher weights to sentences that support the claim, moderate weights to neutral sentences, and lower weights to sentences with minimal relevance to the claim. The darker colour signifies the higher attention weight assigned to the respective sentence from the evidence set and vice-versa.}  
\label{fig:pos_heatmap_1}
\end{figure}
\begin{figure}[]
  \centering
    \includegraphics[width=0.5\textwidth, height=0.36\textheight]{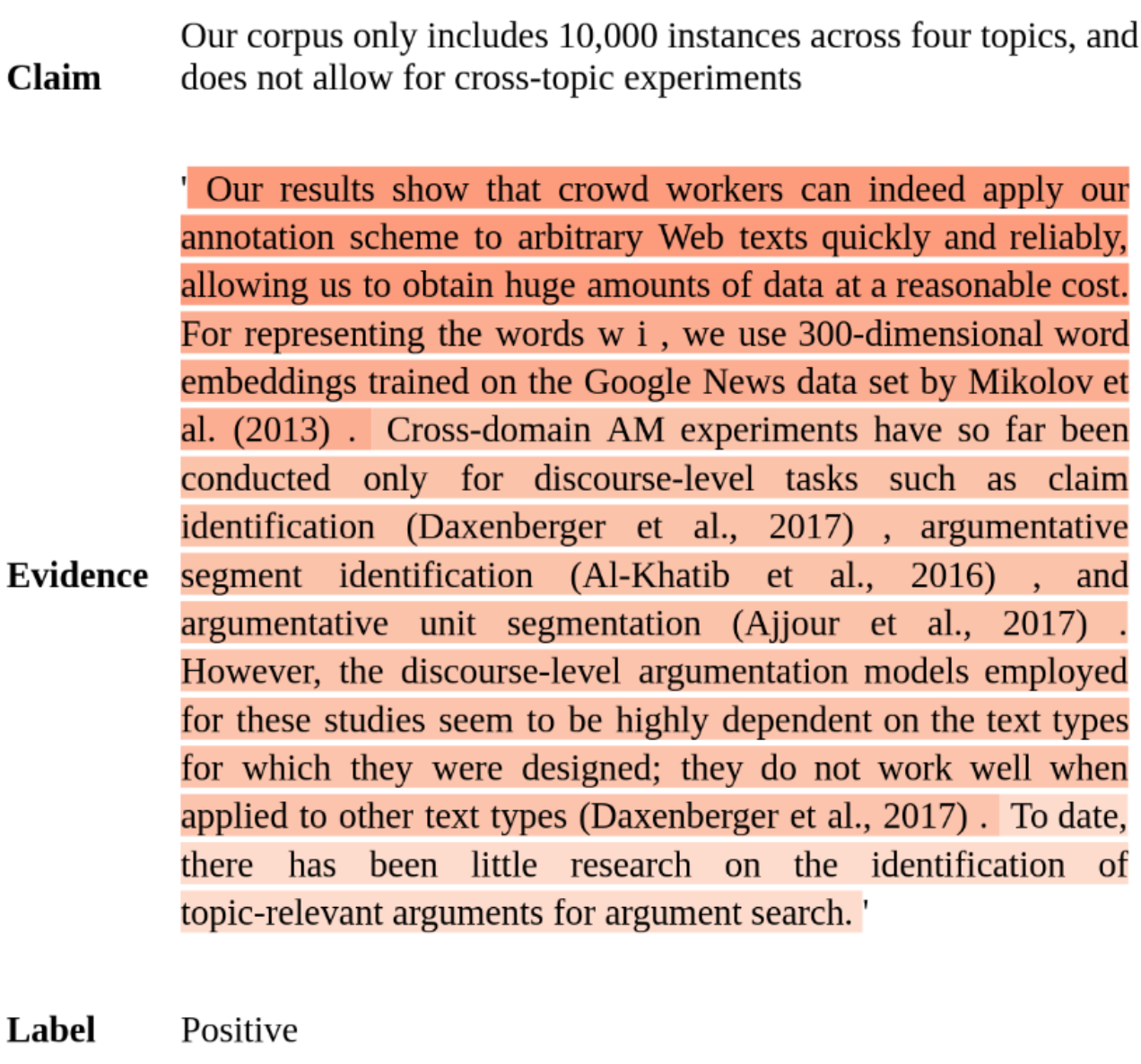}
    \caption{ presents attention heatmaps showing the attention weights between the claim and various sentences of the evidence for the \textit{Supporting} class (positive claims) samples. The heatmaps reveal that the multi-head attention component of the CEM model assigns higher weights to sentences that support the claim, moderate weights to neutral sentences, and lower weights to sentences with minimal relevance to the claim. The darker colour signifies the higher attention weight assigned to the respective sentence from the evidence set and vice-versa.}  
\label{fig:pos_heatmap_2}
\end{figure}
\begin{figure}[]
  \centering
    \includegraphics[width=0.5\textwidth, height=0.48\textheight]{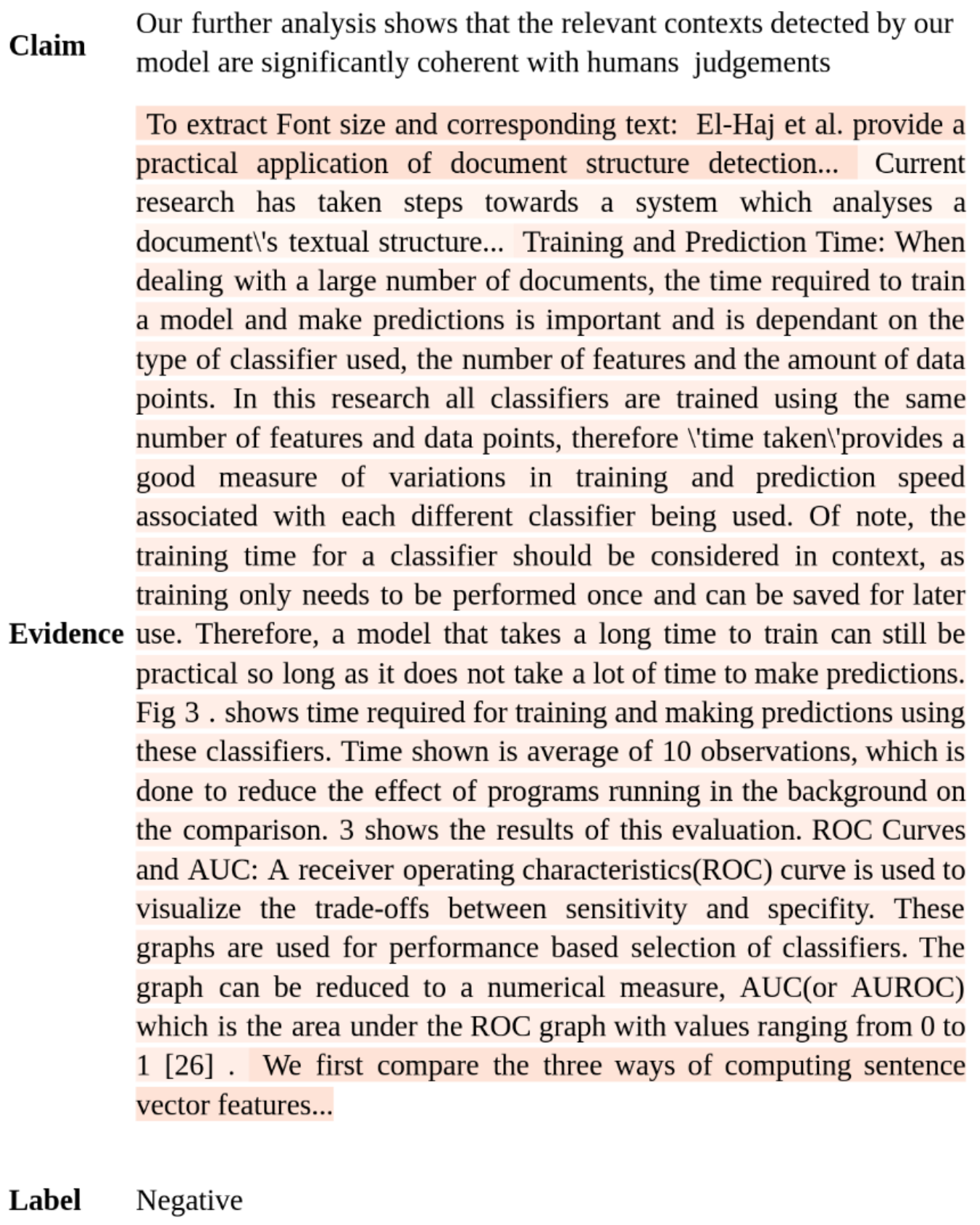}
    \caption{ presents attention heatmaps showing the attention weights between the claim and various sentences of the evidence for the \textit{Refutes} class (negative claims) samples. The heatmaps reveal that the multi-head attention component of the CEM model assigns higher weights to sentences that support the claim, moderate weights to neutral sentences, and lower weights to sentences with minimal relevance to the claim. The darker colour signifies the higher attention weight assigned to the respective sentence from the evidence set and vice-versa.}  
\label{fig:neg_heatmap_1}
\end{figure}

\subsubsection{Error Analysis}
We perform an error analysis on the misclassifications made by baseline models over proposed datasets to evaluate the strengths and weaknesses of the proposed baseline models. We first evaluate the response of baseline models over \textit{SciClaimHunt\_Num} datasets to study the ability of baseline models in scientific claim verification with claims involving cardinal and numeral values. Table~\ref{tab:num_res} presents the performance of the proposed baseline models on the \textit{SciClaimHunt\_Num} datasets, demonstrating the high proficiency of baseline models in verifying scientific claims containing cardinal and numeral values. We manually inspect misclassifications made by the baseline models on the proposed datasets, \textit{SciClaimHunt} and \textit{SciClaimHunt\_Num}, and derive the following key observations: (i) Claims that are only partially supported by the evidence (i.e., where some sentences support the claim while others refute it) pose a challenge, often leading to misclassification by models from support to refutes or refutes to support. (ii) When the evidence discusses the claim without taking a stance for or against it, models incorrectly classify the claim-evidence pair as supported even though it belongs to the refute class. Figure~\ref{fig:pos_heatmap_1},~\ref{fig:pos_heatmap_2},~\ref{fig:neg_heatmap_1} and~\ref{fig:neg_heatmap_2} present heatmaps illustrating the attention weights between the scientific claim and sentences of evidence, as captured by the multi-head attention component of the CEM model. From Figure~\ref{fig:pos_heatmap_1},~\ref{fig:pos_heatmap_2},~\ref{fig:neg_heatmap_1} and~\ref{fig:neg_heatmap_2}, it is evident that the multi-head attention components of the CEM model assign high attention weights to sentences from the evidence set that are highly similar to and support the claim while assigning low attention weights to sentences that are least similar to and refute the claim. From these observations, we can conclude that our proposed baseline model effectively learns the relationship between claims and evidence for scientific claim classification.

\section{Conclusion and Future Work}

This paper presents two novel datasets for scientific claim verification: \textit{SciClaimHunt} and \textit{SciClaimHunt\_Num}, along with several baseline models to assess their effectiveness. We further evaluate the quality, generalizability, and reliability of these datasets through ablation studies, human assessments, and error analyses. Our findings demonstrate that \textit{SciClaimHunt} and \textit{SciClaimHunt\_Num} serve as robust resources for training models in scientific claim verification. Potential future directions for this work include: (i) incorporating scientific claims from low-resource languages, (ii) expanding the dataset to encompass diverse scientific domains such as medicine and disease research, and (iii) integrating images and tabular data from scientific papers as evidence.



\section{Ethics Statement}
\label{sec:ethics}
The SciClaimHunt dataset was created without collecting any personal information. It leverages a publicly available dataset curated by study~\cite{moosavi2021scigen}. The SciClaimHunt dataset is released under the Creative Commons Attribution 4.0 International license (consistent with the original license). The accompanying code is released under the MIT license.

\bibliographystyle{IEEEtran}
\bibliography{IEEEabrv,main_paper}

\begin{thebibliography}{10}
\providecommand{\url}[1]{#1}
\csname url@samestyle\endcsname
\providecommand{\newblock}{\relax}
\providecommand{\bibinfo}[2]{#2}
\providecommand{\BIBentrySTDinterwordspacing}{\spaceskip=0pt\relax}
\providecommand{\BIBentryALTinterwordstretchfactor}{4}
\providecommand{\BIBentryALTinterwordspacing}{\spaceskip=\fontdimen2\font plus
\BIBentryALTinterwordstretchfactor\fontdimen3\font minus \fontdimen4\font\relax}
\providecommand{\BIBforeignlanguage}[2]{{%
\expandafter\ifx\csname l@#1\endcsname\relax
\typeout{** WARNING: IEEEtran.bst: No hyphenation pattern has been}%
\typeout{** loaded for the language `#1'. Using the pattern for}%
\typeout{** the default language instead.}%
\else
\language=\csname l@#1\endcsname
\fi
#2}}
\providecommand{\BIBdecl}{\relax}
\BIBdecl

\bibitem{vlachos2014fact}
A.~Vlachos and S.~Riedel, ``Fact checking: Task definition and dataset construction,'' in \emph{Proceedings of the ACL 2014 workshop on language technologies and computational social science}, 2014, pp. 18--22.

\bibitem{thorne2018fever}
J.~Thorne, A.~Vlachos, C.~Christodoulopoulos, and A.~Mittal, ``Fever: a large-scale dataset for fact extraction and verification,'' in \emph{Proceedings of the 2018 Conference of the North American Chapter of the Association for Computational Linguistics: Human Language Technologies, Volume 1 (Long Papers)}, 2018, pp. 809--819.

\bibitem{hanselowski2019richly}
A.~Hanselowski, C.~Stab, C.~Schulz, Z.~Li, and I.~Gurevych, ``A richly annotated corpus for different tasks in automated fact-checking,'' in \emph{Proceedings of the 23rd Conference on Computational Natural Language Learning (CoNLL)}, 2019, pp. 493--503.

\bibitem{thorne2019fever2}
J.~Thorne, A.~Vlachos, O.~Cocarascu, C.~Christodoulopoulos, and A.~Mittal, ``The fever2. 0 shared task,'' in \emph{Proceedings of the second workshop on Fact Extraction and VERification (FEVER)}, 2019, pp. 1--6.

\bibitem{setty2020deep}
V.~J. Setty and R.~Mishra, ``Deep neural architectures for detecting false claims,'' Oct.~13 2020, uS Patent 10,803,387.

\bibitem{derczynski2017semeval}
L.~Derczynski, K.~Bontcheva, M.~Liakata, R.~Procter, G.~W.~S. Hoi, and A.~Zubiaga, ``Semeval-2017 task 8: Rumoureval: Determining rumour veracity and support for rumours,'' in \emph{Proceedings of the 11th International Workshop on Semantic Evaluation (SemEval-2017)}, 2017, pp. 69--76.

\bibitem{gorrell-etal-2019-semeval}
\BIBentryALTinterwordspacing
G.~Gorrell, E.~Kochkina, M.~Liakata, A.~Aker, A.~Zubiaga, K.~Bontcheva, and L.~Derczynski, ``{S}em{E}val-2019 task 7: {R}umour{E}val, determining rumour veracity and support for rumours,'' in \emph{Proceedings of the 13th International Workshop on Semantic Evaluation}.\hskip 1em plus 0.5em minus 0.4em\relax Minneapolis, Minnesota, USA: Association for Computational Linguistics, Jun. 2019, pp. 845--854. [Online]. Available: \url{https://aclanthology.org/S19-2147}
\BIBentrySTDinterwordspacing

\bibitem{barron2020overview}
A.~Barr{\'o}n-Cedeno, T.~Elsayed, P.~Nakov, G.~Da~San~Martino, M.~Hasanain, R.~Suwaileh, F.~Haouari, N.~Babulkov, B.~Hamdan, A.~Nikolov \emph{et~al.}, ``Overview of checkthat! 2020: Automatic identification and verification of claims in social media,'' in \emph{Experimental IR Meets Multilinguality, Multimodality, and Interaction: 11th International Conference of the CLEF Association, CLEF 2020, Thessaloniki, Greece, September 22--25, 2020, Proceedings 11}.\hskip 1em plus 0.5em minus 0.4em\relax Springer, 2020, pp. 215--236.

\bibitem{elsayed2019checkthat}
T.~Elsayed, P.~Nakov, A.~Barr{\'o}n-Cede{\~n}o, M.~Hasanain, R.~Suwaileh, G.~Da~San~Martino, and P.~Atanasova, ``Checkthat! at clef 2019: Automatic identification and verification of claims,'' in \emph{European Conference on Information Retrieval}, 2019, pp. 309--315.

\bibitem{nakov2018overview}
P.~Nakov, A.~Barr{\'o}n-Cedeno, T.~Elsayed, R.~Suwaileh, L.~M{\`a}rquez, W.~Zaghouani, P.~Atanasova, S.~Kyuchukov, and G.~Da~San~Martino, ``Overview of the clef-2018 checkthat! lab on automatic identification and verification of political claims,'' in \emph{Experimental IR Meets Multilinguality, Multimodality, and Interaction: 9th International Conference of the CLEF Association, CLEF 2018, Avignon, France, September 10-14, 2018, Proceedings 9}.\hskip 1em plus 0.5em minus 0.4em\relax Springer, 2018, pp. 372--387.

\bibitem{hassan2017claimbuster}
N.~Hassan, G.~Zhang, F.~Arslan, J.~Caraballo, D.~Jimenez, S.~Gawsane, S.~Hasan, M.~Joseph, A.~Kulkarni, A.~K. Nayak \emph{et~al.}, ``Claimbuster: The first-ever end-to-end fact-checking system,'' \emph{Proceedings of the VLDB Endowment}, vol.~10, no.~12, pp. 1945--1948, 2017.

\bibitem{SADHAN_mishra}
\BIBentryALTinterwordspacing
R.~Mishra and V.~Setty, ``Sadhan: Hierarchical attention networks to learn latent aspect embeddings for fake news detection,'' in \emph{Proceedings of the 2019 ACM SIGIR International Conference on Theory of Information Retrieval}, ser. ICTIR '19.\hskip 1em plus 0.5em minus 0.4em\relax New York, NY, USA: Association for Computing Machinery, 2019, p. 197–204. [Online]. Available: \url{https://doi.org/10.1145/3341981.3344229}
\BIBentrySTDinterwordspacing

\bibitem{mishra-etal-2020-generating}
\BIBentryALTinterwordspacing
R.~Mishra, D.~Gupta, and M.~Leippold, ``Generating fact checking summaries for web claims,'' in \emph{Proceedings of the Sixth Workshop on Noisy User-generated Text (W-NUT 2020)}, W.~Xu, A.~Ritter, T.~Baldwin, and A.~Rahimi, Eds.\hskip 1em plus 0.5em minus 0.4em\relax Online: Association for Computational Linguistics, Nov. 2020, pp. 81--90. [Online]. Available: \url{https://aclanthology.org/2020.wnut-1.12}
\BIBentrySTDinterwordspacing

\bibitem{wadden2020fact}
D.~Wadden, S.~Lin, K.~Lo, L.~L. Wang, M.~van Zuylen, A.~Cohan, and H.~Hajishirzi, ``Fact or fiction: Verifying scientific claims,'' in \emph{Proceedings of the 2020 Conference on Empirical Methods in Natural Language Processing (EMNLP)}, 2020, pp. 7534--7550.

\bibitem{wadden2022scifact}
D.~Wadden, K.~Lo, B.~Kuehl, A.~Cohan, I.~Beltagy, L.~L. Wang, and H.~Hajishirzi, ``Scifact-open: Towards open-domain scientific claim verification,'' in \emph{Findings of the Association for Computational Linguistics: EMNLP 2022}, 2022, pp. 4719--4734.

\bibitem{wright2022generating}
D.~Wright, D.~Wadden, K.~Lo, B.~Kuehl, A.~Cohan, I.~Augenstein, and L.~L. Wang, ``Generating scientific claims for zero-shot scientific fact checking,'' in \emph{Proceedings of the 60th Annual Meeting of the Association for Computational Linguistics (Volume 1: Long Papers)}, 2022, pp. 2448--2460.

\bibitem{vladika2023scientific}
J.~Vladika and F.~Matthes, ``Scientific fact-checking: A survey of resources and approaches,'' in \emph{Findings of the Association for Computational Linguistics: ACL 2023}, 2023, pp. 6215--6230.

\bibitem{guo2022survey}
Z.~Guo, M.~Schlichtkrull, and A.~Vlachos, ``A survey on automated fact-checking,'' \emph{Transactions of the Association for Computational Linguistics}, vol.~10, pp. 178--206, 2022.

\bibitem{popat2017truth}
K.~Popat, S.~Mukherjee, J.~Str{\"o}tgen, and G.~Weikum, ``Where the truth lies: Explaining the credibility of emerging claims on the web and social media,'' in \emph{Proceedings of the 26th international conference on world wide web companion}, 2017, pp. 1003--1012.

\bibitem{thorne2018fact}
J.~Thorne, A.~Vlachos, O.~Cocarascu, C.~Christodoulopoulos, and A.~Mittal, ``The fact extraction and verification (fever) shared task,'' in \emph{Proceedings of the First Workshop on Fact Extraction and VERification (FEVER)}, 2018, pp. 1--9.

\bibitem{alhindi2018your}
T.~Alhindi, S.~Petridis, and S.~Muresan, ``Where is your evidence: Improving fact-checking by justification modeling,'' in \emph{Proceedings of the first workshop on fact extraction and verification (FEVER)}, 2018, pp. 85--90.

\bibitem{chen2019seeing}
S.~Chen, D.~Khashabi, W.~Yin, C.~Callison-Burch, and D.~Roth, ``Seeing things from a different angle: Discovering diverse perspectives about claims,'' in \emph{Proceedings of the 2019 Conference of the North American Chapter of the Association for Computational Linguistics: Human Language Technologies, Volume 1 (Long and Short Papers)}, 2019, pp. 542--557.

\bibitem{augenstein2019multifc}
I.~Augenstein, C.~Lioma, D.~Wang, L.~C. Lima, C.~Hansen, C.~Hansen, and J.~G. Simonsen, ``Multifc: A real-world multi-domain dataset for evidence-based fact checking of claims,'' in \emph{Proceedings of the 2019 Conference on Empirical Methods in Natural Language Processing and the 9th International Joint Conference on Natural Language Processing (EMNLP-IJCNLP)}, 2019, pp. 4685--4697.

\bibitem{atanasova2024multi}
P.~Atanasova, ``Multi-hop fact checking of political claims,'' in \emph{Accountable and Explainable Methods for Complex Reasoning over Text}.\hskip 1em plus 0.5em minus 0.4em\relax Springer, 2024, pp. 131--151.

\bibitem{sathe2020automated}
A.~Sathe, S.~Ather, T.~M. Le, N.~Perry, and J.~Park, ``Automated fact-checking of claims from wikipedia,'' in \emph{Proceedings of the Twelfth Language Resources and Evaluation Conference}, 2020, pp. 6874--6882.

\bibitem{saakyan2021covid}
A.~Saakyan, T.~Chakrabarty, and S.~Muresan, ``Covid-fact: Fact extraction and verification of real-world claims on covid-19 pandemic,'' in \emph{Proceedings of the 59th Annual Meeting of the Association for Computational Linguistics and the 11th International Joint Conference on Natural Language Processing (Volume 1: Long Papers)}, 2021, pp. 2116--2129.

\bibitem{schuster2021get}
T.~Schuster, A.~Fisch, and R.~Barzilay, ``Get your vitamin c! robust fact verification with contrastive evidence,'' in \emph{Proceedings of the 2021 Conference of the North American Chapter of the Association for Computational Linguistics: Human Language Technologies}, 2021, pp. 624--643.

\bibitem{aly2021feverous}
R.~Aly, Z.~Guo, M.~Schlichtkrull, J.~Thorne, A.~Vlachos, C.~Christodoulopoulos, O.~Cocarascu, and A.~Mittal, ``Feverous: Fact extraction and verification over unstructured and structured information,'' in \emph{35th Conference on Neural Information Processing Systems, NeurIPS 2021}.\hskip 1em plus 0.5em minus 0.4em\relax Neural Information Processing Systems foundation, 2021.

\bibitem{moosavi2021scigen}
N.~Moosavi, A.~R{\"u}ckl{\'e}, D.~Roth, and I.~Gurevych, ``Scigen: a dataset for reasoning-aware text generation from scientific tables,'' in \emph{Proceedings of the Neural Information Processing Systems Track on Datasets and Benchmarks 1 (NeurIPS Datasets and Benchmarks 2021)}.\hskip 1em plus 0.5em minus 0.4em\relax Neural Information Processing Systems Foundation, 2021.

\bibitem{daipromptagator}
Z.~Dai, V.~Y. Zhao, J.~Ma, Y.~Luan, J.~Ni, J.~Lu, A.~Bakalov, K.~Guu, K.~Hall, and M.-W. Chang, ``Promptagator: Few-shot dense retrieval from 8 examples,'' in \emph{The Eleventh International Conference on Learning Representations}, 2022.

\bibitem{touvron2023llama}
H.~Touvron, T.~Lavril, G.~Izacard, X.~Martinet, M.-A. Lachaux, T.~Lacroix, B.~Rozi{\`e}re, N.~Goyal, E.~Hambro, F.~Azhar \emph{et~al.}, ``Llama: Open and efficient foundation language models,'' \emph{arXiv preprint arXiv:2302.13971}, 2023.

\bibitem{krippendorff2011computing}
K.~Krippendorff, ``Computing krippendorff’s alpha-reliability,'' 2011.

\bibitem{fleiss1971measuring}
J.~L. Fleiss, ``Measuring nominal scale agreement among many raters.'' \emph{Psychological bulletin}, vol.~76, no.~5, p. 378, 1971.

\bibitem{reimers-2019-sentence-bert}
\BIBentryALTinterwordspacing
N.~Reimers and I.~Gurevych, ``Sentence-bert: Sentence embeddings using siamese bert-networks,'' in \emph{Proceedings of the 2019 Conference on Empirical Methods in Natural Language Processing}.\hskip 1em plus 0.5em minus 0.4em\relax Association for Computational Linguistics, 11 2019. [Online]. Available: \url{https://arxiv.org/abs/1908.10084}
\BIBentrySTDinterwordspacing

\bibitem{lewis2020retrieval}
P.~Lewis, E.~Perez, A.~Piktus, F.~Petroni, V.~Karpukhin, N.~Goyal, H.~K{\"u}ttler, M.~Lewis, W.-t. Yih, T.~Rockt{\"a}schel \emph{et~al.}, ``Retrieval-augmented generation for knowledge-intensive nlp tasks,'' \emph{Advances in Neural Information Processing Systems}, vol.~33, pp. 9459--9474, 2020.

\bibitem{team2024gemma}
G.~Team, M.~Riviere, S.~Pathak, P.~G. Sessa, C.~Hardin, S.~Bhupatiraju, L.~Hussenot, T.~Mesnard, B.~Shahriari, A.~Ram{\'e} \emph{et~al.}, ``Gemma 2: Improving open language models at a practical size,'' \emph{arXiv preprint arXiv:2408.00118}, 2024.

\bibitem{vaswani2017attention}
A.~Vaswani, ``Attention is all you need,'' \emph{Advances in Neural Information Processing Systems}, 2017.

\bibitem{nikolentzos2020message}
G.~Nikolentzos, A.~Tixier, and M.~Vazirgiannis, ``Message passing attention networks for document understanding,'' in \emph{Proceedings of the aaai conference on artificial intelligence}, vol.~34, no.~05, 2020, pp. 8544--8551.

\bibitem{zhang2020every}
Y.~Zhang, X.~Yu, Z.~Cui, S.~Wu, Z.~Wen, and L.~Wang, ``Every document owns its structure: Inductive text classification via graph neural networks,'' in \emph{Proceedings of the 58th Annual Meeting of the Association for Computational Linguistics}, 2020, pp. 334--339.

\bibitem{velickovic2018graph}
\BIBentryALTinterwordspacing
P.~Veli{\v{c}}kovi{\'{c}}, G.~Cucurull, A.~Casanova, A.~Romero, P.~Li{\`{o}}, and Y.~Bengio, ``{Graph Attention Networks},'' \emph{International Conference on Learning Representations}, 2018. [Online]. Available: \url{https://openreview.net/forum?id=rJXMpikCZ}
\BIBentrySTDinterwordspacing

\end{thebibliography}

\end{document}